\documentclass[runningheads]{llncs}
\usepackage[T1]{fontenc}
\usepackage{graphicx}
\usepackage{booktabs}
\usepackage{makecell}
\usepackage{amsmath}
\usepackage{multirow}
\usepackage{xcolor}
\usepackage{subcaption}
\begin{document}
\title{Minimum Data, Maximum Impact: 20 annotated samples for explainable lung nodule classification}
\titlerunning{Minimum Data, Maximum Impact}
% \title{Contribution Title}
\author{
Luisa Gallée \inst{1,2,3}\orcidID{0000-0001-5556-7395}\and
Catharina Silvia Lisson \inst{2}\orcidID{0009-0000-5668-3216}\and
Christoph Gerhard Lisson \inst{2}\and
Daniela Drees \inst{2}\and
Felix Weig \inst{2}\and
Daniel Vogele \inst{2}\orcidID{0000-0002-6421-1400}\and
Meinrad Beer \inst{2,3}\orcidID{0000-0001-7523-1979}\and
Michael Götz \inst{1,2,3}\orcidID{0000-0003-0984-224X}
% First Author\inst{1}\orcidID{0000-1111-2222-3333} \and
% Third Author\inst{3}\orcidID{2222--3333-4444-5555}
}
\authorrunning{L. Gallée et al.}
\institute{
Experimental Radiology, Ulm University Medical Center, Ulm, Germany \and
Department of Diagnostic and Interventional Radiology, Ulm University Medical Center, Ulm, Germany \and
XAIRAD - Cooperation for Artificial Intelligence in Experimental Radiology, Ulm, Germany % Princeton University, Princeton NJ 08544, USA \and
}
\maketitle              % typeset the header of the contribution
\begin{abstract}
Classification models that provide human‑interpretable explanations enhance clinicians’ trust and usability in medical image diagnosis.
One research focus is the integration and prediction of pathology-related visual attributes used by radiologists alongside the diagnosis, aligning AI decision-making with clinical reasoning.
Radiologists use attributes like shape and texture as established diagnostic criteria and mirroring these in AI decision‑making both enhances transparency and enables explicit validation of model outputs.
However, the adoption of such models is limited by the scarcity of large-scale medical image datasets annotated with these attributes.
To address this challenge, we propose synthesizing attribute-annotated data using a generative model.
We enhance the Diffusion Model with attribute conditioning and train it using only 20 attribute-labeled lung nodule samples from the LIDC-IDRI dataset.
Incorporating its generated images into the training of an explainable model boosts performance, increasing attribute prediction accuracy by 13.4\% and target prediction accuracy by 1.8\% compared to training with only the small real attribute-annotated dataset.
This work highlights the potential of synthetic data to overcome dataset limitations, enhancing the applicability of explainable models in medical image analysis.

% The abstract should briefly summarize the contents of the paper in

% \keywords{First keyword  \and Second keyword \and Another keyword.}
\end{abstract}

\section{Introduction}
When working with a medical dataset containing diagnosis labels, such as malignant or benign tumors, a straightforward approach is to train a classification model to predict tumor malignancy. 
However, for physicians, merely predicting a disease is insufficient to establish trust in the model and consequently adopt it in clinical practice. Instead, a more trustworthy approach involves integrating the same diagnostic criteria that radiologists use when making decisions into the model.
\emph{Why does the model classify the tumor as malignant? Which criteria does it consider fulfilled?}
To facilitate such discussion, the model’s training dataset must include annotations for these decision criteria alongside the disease labels. In radiological diagnosis, these criteria correspond to visually identifiable attributes of the pathology, including appearance characteristics such as shape and texture.
However, obtaining such annotations requires additional effort, as they are not routinely recorded in standard radiology reports. As a result, large-scale datasets with these annotations are scarce, limiting their applicability for explainable deep learning models.

\begin{figure}[!htbp]
\centering
\includegraphics[width=0.7\textwidth]{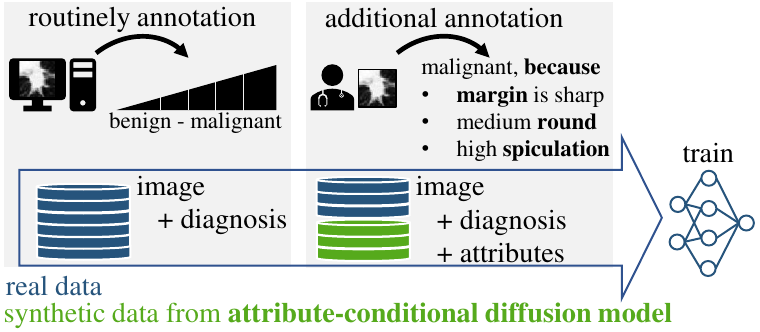}
\caption{Large clinical datasets with disease labels are feasible to extract from routine data. However, diagnostic attributes require extra annotation, limiting dataset size. Synthetic augmentation can expand annotated datasets, enabling models to learn attribute-based reasoning in disease prediction.
} \label{fig:intro}
\end{figure}

In this paper, we explore a method for handling such incomplete datasets, using the LIDC-IDRI dataset \cite{armato_iii_lung_data_2015} to classify the malignancy of lung nodules while incorporating their visual characteristics as reasoning attributes.
We evaluate a two-step approach: first, augmenting medical images using a generative model conditioned on the attributes; second, adding these synthetic images to the training of explainable image classification models, thereby mitigating the scarcity of human-annotated images, as depicted in Figure \ref{fig:intro}.

The generation of medical images has been explored in research for various motivations, including reducing the cost and time required for data acquisition, enabling digital twin technology, handling diverse patient populations, representing multimodal datasets, and ensuring patient data privacy  \cite{pezoulas2024synthetic,giuffre2023harnessing,kazeminia2020gans,dumont2021overcoming,zhou2024privacy}.
Related work most relevant to our task, focusing on generative AI for lung nodule images, utilizes some nodule characteristics such as size \cite{NISHIO2020104032} as conditioning criteria to improve the quality of the generated images \cite{HAVAEI2021102106}. Research on using generated images for training classification models, both with attribute conditioning \cite{Wang2021Realistic,toda2021synthetic} and without \cite{Onishi2020Multiplanar,Onishi2020Investigation}, demonstrates performance benefits when these images are used as pre-training data. While some attributes of lung nodules have been incorporated into generative AI models, their use has been limited to improving image quality for disease classification, without providing any insight into the diagnostic reasoning behind the predictions.

Our focus is on investigating how recent advancements in generative models can support the development of explainable AI algorithms.
We generate images that satisfy combinations of multiple non-binary attributes and use them to train explainable downstream classification models. 
The explainability of these models includes both target prediction and attribute prediction.

Methodologically, we employ a Diffusion Model for image synthesis \cite{ho2020denoising}. Compared to Generative Adversarial Networks (GANs) and Variational Autoencoders (VAEs), Diffusion Models have demonstrated superior robustness and performance \cite{goodfellow2014generative,kingma2013auto,zhan2024conditional}. The conditioning capabilities of Diffusion Models were demonstrated by Rombach et al. \cite{rombach2022high}, who introduced cross-attention mechanisms for conditioning on various modalities. 
Prior research on multi-label conditioning in Diffusion Models has primarily focused on binary \cite{lisanti2024conditioning} or nominal \cite{kong2023leveraging} attributes in general-domain images.
The novelties presented in this work are summarized as follows:
\begin{itemize}
\item \textbf{Attribute-Conditional Generative AI}: We present a Diffusion Model with conditioning on comprehensive diagnostic criteria (attributes) for medical image synthesis.
\item \textbf{Semi-Conditional Training}: Using labeled and unlabeled data to train the Diffusion Model improves the quality of generated images.
\item \textbf{Substituting Sparse Datasets}: We leverage the generated annotated data to train SOTA explainable classification models, improving performance on a sparsely annotated subset of the LIDC-IDRI medical benchmark dataset. %Enhancing, Enlarging
\end{itemize}

The code is publicly available at \url{https://github.com/XRad-Ulm/Attribute-Conditional-Diffusion-Model}.

\section{Methods}
% The proposed solution follows a two-step approach: first, augmenting a small attribute-labeled dataset using the proposed \textit{Attribute-Conditional} \textit{Diffusion}  \textit{Model}, and second, integrating the synthetic data into the training of SOTA attribute-based explainable classifiers.

\subsection{Attribute-Conditional Diffusion Model}
For this work, we employ an attribute-conditional Diffusion Model that processes an image alongside conditioning inputs in the form of multiple metric attributes. The model generates new images that satisfy specific attribute combinations, allowing for the synthesis of an attribute-annotated dataset.
The architecture builds upon an existing unconditional Diffusion Model introduced by Ho et al. \cite{ho2020denoising}. We utilize a convolution-based U-Net, structured with two encoder blocks, a central bottleneck, and two decoder blocks. Each block incorporates residual bottlenecks, following the design described by Ma et al. \cite{ma2018shufflenet,guocheng2022MNISTDiffusion}.

The conditioning mechanism is applied within the network’s bottleneck.
The attribute vector comprising a single continuous value in the range $[0,1]$ for each attribute (margin, sphericity, spiculation, etc.) is first projected onto the bottleneck dimensionality via a linear transformation.
This transformed attribute latent vector is then integrated with the image data’s latent representation through a cross-attention mechanism \cite{vaswani2017attention}, employing a single-token cross-attention approach without residual connections to reduce parameter count. 

\begin{figure}[!htbp]
\centering
\includegraphics[width=0.8\textwidth]{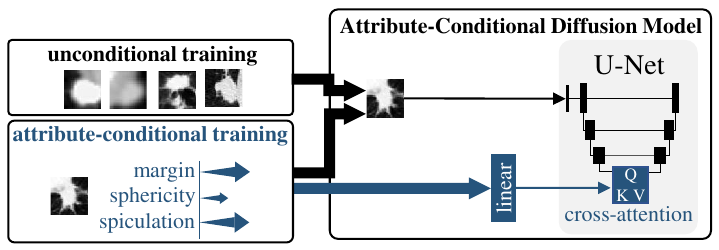}
\caption{\textbf{Generative Model} A Diffusion Model for synthesis of attribute-labeled images. Semi-conditional training incorporates both labeled and unlabeled samples by dynamically activating or deactivating cross-attention.
} \label{fig:vision}
\end{figure}

\paragraph{Semi-Conditional Training}
The cross-attention mechanism provides flexibility for semi-conditional training (see Figure \ref{fig:vision}). 
On one hand, attribute conditioning is achieved by passing the latent vector from cross-attention to the decoder. 
On the other hand, samples without attribute labels are processed without the attention; their latent representations are directly passed to the decoder. This semi-conditional learning approach is motivated by the limited availability of attribute-labeled samples, which alone may be insufficient for Diffusion Models to generate high-quality images. Unlabeled images, which are relatively easy to obtain in clinical routine, can effectively contribute to this process.

\subsection{Attribute-Explainable Classification Model}
To evaluate the effect of our synthetic dataset on explainable AI performance, we train two hierarchical classifiers with it, HierViT \cite{GalleeHierViT2025} and the Concept Bottleneck Model \cite{koh2020concept}. Both architectures first predict visual attributes and then use those attributes to classify the target. HierViT further enhances interpretability by displaying prototypes, representative samples that illustrate each attribute prediction, and generating attribute attention heatmaps. We selected these models because they achieve state-of-the-art accuracy in both attribute and target prediction and offer comprehensive, human interpretable explanations, which have been shown to foster trust among medical professionals \cite{gallee2024Evaluating}.

\section{Experiments}
\subsection{Dataset}
The Lung Image Database Consortium and Image Database Resource Initiative (CC BY 3.0) \cite{armato_iii_lung_data_2015} provides a densely annotated CT dataset of patients with non-small cell lung cancer. Up to four radiologists segmented lung nodules and rated malignancy on a scale from 1 to 5 \cite{armato_iii_lung_2011}. 
Additionally, the attributes, namely subtlety, internal structure, calcification, sphericity, margin, lobulation, spiculation, and texture, are also annotated using multi-level rating scales. The attributes are well-established diagnostic criteria for assessing the malignancy of lung nodules.

Preprocessing is performed in the same manner as in the comparative experiments described in Gallée et al. \cite{GalleeHierViT2025}, which involves excluding nodules identified by fewer than three radiologists or those smaller than 3 mm. Cropouts are generated using the smallest square bounding box and resized to 32\(\times\)32 pixels for the Diffusion Model, and to 224\(\times\)224 pixels for the HierViT model using \texttt{pylidc} \cite{LIDC_hancock}.
The dataset consists of 27,379 samples and is evaluated using 5-fold stratified cross-validation by patient, with 10\% of the training data reserved for validation.

To reflect our application scenario of sparsely attribute annotated data, only a subset $r$ from the real LIDC IDRI dataset contains additional attribute annotations and is used to train the generative model. In semi-conditional mode the generative model also leverages all remaining images without annotations.
For subsequent classification model training, the synthetic attribute-annotated data $s$ is additionally included.

\subsection{Implementation Details}
\textit{Training of Generative Model }
Training of the semi-conditional model until loss convergence after 100 epochs on a GeForce RTX 3090 graphics card yielded an average runtime of 40 minutes.
For the training of the classification model synthetic samples were generated by combining real attribute values. Preliminary studies showed that sampling images with random, unrealistic attribute combinations led to poor image quality.
The generated samples are also annotated with the respective ground truth malignancy score, as implied by the attribute combination.

\textit{Training Classification Model }
The model and training setup of HierViT are adopted from the original work \cite{GalleeHierViT2025}. 
For the Concept Bottleneck Model (CB), we adopt the 'joint' configuration to tightly integrate attribute estimation and target classification. The architecture comprises four fully connected layers with output sizes \texttt{[256, 128, 8, 5]}. The third fc-layer generates predictions for the eight visual attributes, and the final layer produces classification scores for the five target classes.

Semi-conditional attribute learning is facilitated by selectively activating or deactivating the attribute loss term, depending on whether the samples contain attribute labels. The real dataset and the synthetic dataset are randomly shuffled for training.

Neither the Attribute-Conditional Diffusion Model nor the classification models were pretrained.

\section{Results}

\subsection{Image Synthesis}
\paragraph{User Study}
\begin{figure}[!htbp]
\centering
\includegraphics[width=0.84\textwidth]{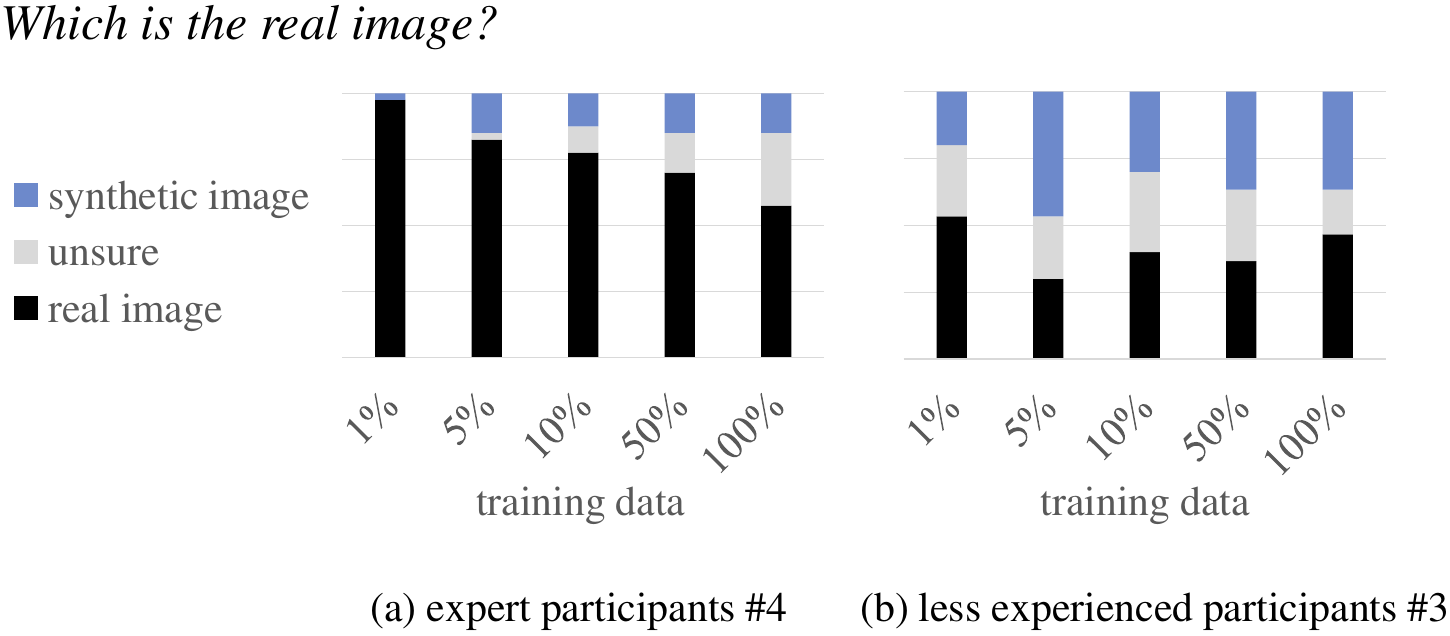}
\caption{Results of the user study assessing the realism of generated images. The plots show the choice made by (a) experts and (b) less experienced participants when asked to identify the real image in a pair. The x-axis denotes the proportion of training data used to train the Attribute-Conditional Diffusion Model.}
\label{results_us}
\end{figure}
To determine whether experts could distinguish real from synthetic images, we asked participants to view fifty image pairs. Each pair contained one real image and one generated image and participants indicated which image they believed to be real. The generated images were produced by the attention-conditional diffusion models trained on different data volumes. The survey imposed no time limits and did not record response times. Seven domain experts from the University Hospital of Ulm including four senior radiologists with over ten years of experience and three less experienced clinicians participated. Experts identified real images more reliably when generation quality was low but their accuracy declined as quality improved. Less experienced participants showed similarly low discrimination performance regardless of the amount of training data. These results suggest that experienced radiologists can detect model shortcomings when training data are limited yet struggle to distinguish real from synthetic images once image quality improves.

\paragraph{Quantitative Results}
\begin{figure}[!htbp]
\centering
\includegraphics[width=0.75\textwidth]{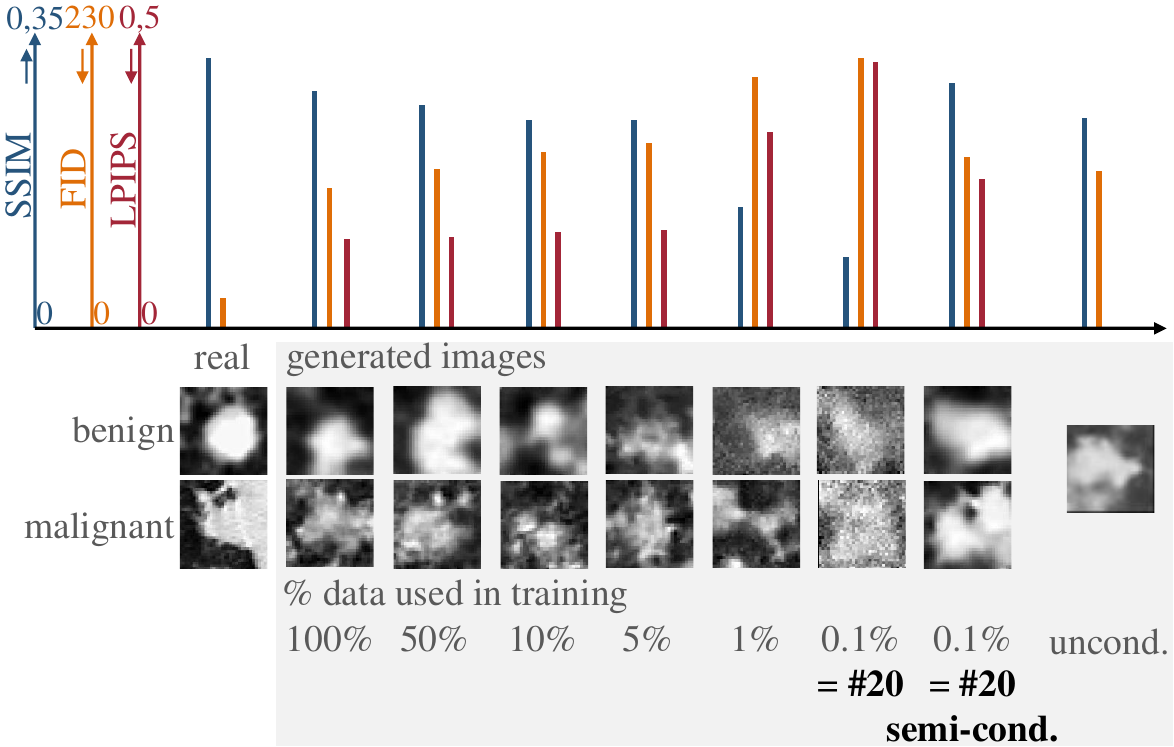}
\caption{Qualitative and quantitative assessment of generated data: Structural Similarity Index Measure (SSIM), Fréchet inception distance (FID), Learned Perceptual Image Patch Similarity (LPIPS). The baseline FID is established by splitting the real test set equally into “real” and “fake” distributions.} \label{fig:res_quant}
\end{figure}
Figure \ref{fig:res_quant} presents examples and quantitative metrics. As attribute labeled training data decreases, SSIM declines while FID and LPIPS increase, reflecting the model’s high data dependency. 
Training with only 20 annotated images yields noisy outputs. Adding semi-conditional training with the remaining images as unconditional data restores SSIM to levels seen with 50-100\% conditional data and brings FID and LPIPS into the range of 1-50\% conditional data.
Training without any attribute information lags behind attribute-conditioned training, underscoring the importance of attribute guidance.

\subsection{Classification Performance}

Table \ref{tab:classperformance} reports classification model performance with synthetic images incorporated into the training set.
The respective first rows represent the straightforward approach of training the classification models using only the limited attribute-annotated real samples and no synthetic data  ($r=20$, $s=0$).
As expected, both HierViT and CB exhibit low attribute prediction accuracy with high standard deviations. However, HierViT outperforms CB in attribute prediction, likely due to its prototype-based learning approach, which enhances stability under conditions of limited training data \cite{snell2017prototypical}.
Remarkably, the target accuracy of HierViT is lower compared to the scenario with high attribute availability (last row), despite the use of the same large number of target labels.
This suggests that insufficient attribute-annotated data negatively impacts target prediction performance of HierViT.
Expanding the training dataset with $s=2000$ attribute-labeled synthetic samples, as seen in rows two and three, leads to an improvement in HierViT's performance despite the limited availability of real attribute-labeled training data. Semi-conditional training of the Diffusion Model further enhances performance. For CB, both attribute and target prediction performances exhibit high variability and inconsistent results, indicating limited robustness of the model.
Further ablation studies on HierViT evaluate the impact of more real and synthetic data ratios on model performance.
With $r=200, s=2000$ samples, performance ($93.2\,\%$ mean attribute, $92.8\,\%$ target) remains comparable to using $r=20$ but at a higher annotation cost.
Using $r=20, s=20'000$ samples lowers performance ($91.9\,\%$ mean attribute, $90.3\,\%$ target), suggesting that excessive synthetic data may introduce noise and reduce classification accuracy.

\begin{table*}[!htbp]
  \caption{\textbf{Classification Performance} 
  Results are reported in the Within-1-Accuracy metric (\%) in mean (black) and standard deviation (gray). A 95\% binomial confidence interval is provided for the attribute mean and target.
  Semi-conditional training of the Diffusion Model is marked with an asterisk *.
  }
  \label{tab:classperformance}
  \centering
  \fontsize{8}{9}\selectfont
  \begin{tabular}{ccc|ccccccccl|l}
    \toprule
    \multicolumn{2}{c}{attribute data}
    &&\multicolumn{9}{c}{attributes} & \multicolumn{1}{c}{target}\\
    \cline{4-12}
    \cline{13-13}
    \multicolumn{1}{c}{\textit{r}}&\multicolumn{1}{c}{\textit{s}}&& sub & is & cal & sph & mar & lob & spic & tex & \multicolumn{1}{c}{\text{\O}} & \multicolumn{1}{c}{malignancy}\\
    \cline{1-13}
    \multicolumn{2}{l}{\textbf{HierViT \cite{GalleeHierViT2025}}}&&&&&&&&&&\\
    \multirow{2}{*}{\makecell[l]{20}}&\multirow{2}{*}{\makecell[l]{0}}&&85.1&99.8&74.2&79.0&80.3&80.0&85.1&57.3
    &80.1 [79.6,80.6]
    &90.9 [90.6,91.2]
    \\
    &&&\textcolor{gray}{13.3}&\textcolor{gray}{0.3}&\textcolor{gray}{32.4}&\textcolor{gray}{19.3}&\textcolor{gray}{11.6}&\textcolor{gray}{15.4}&\textcolor{gray}{5.4}&\textcolor{gray}{41.4}&&\textcolor{gray}{3.6}\\
    \multirow{2}{*}{\makecell[l]{20}}&\multirow{2}{*}{\makecell[l]{2000}}&&94.2&99.4&91.3&96.2&89.7&90.4&88.8&92.5
    &92.9 [92.6,93.2]
    &92.0 [91.7,92.3]
    \\
    &&&\textcolor{gray}{4.1}&\textcolor{gray}{0.3}&\textcolor{gray}{5.5}&\textcolor{gray}{1.7}&\textcolor{gray}{2.9}&\textcolor{gray}{3.6}&\textcolor{gray}{2.7}&\textcolor{gray}{2.4}&&\textcolor{gray}{1.3}\\    
    \multirow{2}{*}{\makecell[l]{20*}}&\multirow{2}{*}{\makecell[l]{2000}}&&97.0&99.8&90.0&97.2&91.3&91.0&89.2&92.5
    &93.5 [93.2,93.8]
    &92.7 [92.4,93.0]
    \\
    &&&\textcolor{gray}{0.8}&\textcolor{gray}{0.3}&\textcolor{gray}{5.4}&\textcolor{gray}{1.1}&\textcolor{gray}{1.7}&\textcolor{gray}{4.1}&\textcolor{gray}{0.8}&\textcolor{gray}{1.4}&&\textcolor{gray}{1.6}\\  
    \multirow{2}{*}{\makecell[l]{19713}}&\multirow{2}{*}{\makecell[l]{0}}&&96.3&99.8&95.5&97.4&92.7&94.3&90.8&93.3
    &95.0 [94.7,95.3]
    &94.8 [94.5,95.1]
    \\
    &&&\textcolor{gray}{0.9}&\textcolor{gray}{0.3}&\textcolor{gray}{2.1}&\textcolor{gray}{0.8}&\textcolor{gray}{1.7}&\textcolor{gray}{2.9}&\textcolor{gray}{1.4}&\textcolor{gray}{1.4}&&\textcolor{gray}{1.4}\\
    \cline{1-13}
    \multicolumn{2}{l}{\textbf{CB \cite{koh2020concept}}}&&&&&&&&&&\\
    20&0&&10.2&20.2&24.1&17.1&25.7&25.7&12.0&18.0&19.1 [18.6,19.6]&91.4 [91.1,91.7]\\
    &&&\textcolor{gray}{5.1}&\textcolor{gray}{7.0}&\textcolor{gray}{8.9}&\textcolor{gray}{12.4}&\textcolor{gray}{9.8}&\textcolor{gray}{10.5}&\textcolor{gray}{7.5}&\textcolor{gray}{7.4}&&\textcolor{gray}{1.4}\\
    20&2000&&20.4&28.7&26.7&34.0&12.4&37.6&29.2&26.7&27.0 [26.5,27.5]&77.2 [76.7,77.7]\\
    &&&\textcolor{gray}{39.1}&\textcolor{gray}{34.7}&\textcolor{gray}{34.2}&\textcolor{gray}{43.4}&\textcolor{gray}{25.9}&\textcolor{gray}{33.9}&\textcolor{gray}{33.5}&\textcolor{gray}{38.9}&&\textcolor{gray}{3.8}\\
    
    20*&2000&&42.5&69.9&33.3&55.9&41.2&53.6&49.0&43.9&49.5 [48.9,50.1]&78.1 [77.6,78.6]\\
    &&&\textcolor{gray}{39.2}&\textcolor{gray}{36.0}&\textcolor{gray}{33.2}&\textcolor{gray}{31.6}&\textcolor{gray}{33.1}&\textcolor{gray}{33.9}&\textcolor{gray}{35.3}&\textcolor{gray}{32.9}&&\textcolor{gray}{10.4}\\
    19713&0&&94.7&99.0&93.9&97.0&94.6&89.0&92.4&92.6&94.2 [93.3,94.7]&91.6 [91.3,91.9]\\
    &&&\textcolor{gray}{1.5}&\textcolor{gray}{1.5}&\textcolor{gray}{2.3}&\textcolor{gray}{2.5}&\textcolor{gray}{0.9}&\textcolor{gray}{2.3}&\textcolor{gray}{3.8}&\textcolor{gray}{1.0}&&\textcolor{gray}{1.6}\\
    \bottomrule
  \end{tabular}
\end{table*}

\section{Discussion and Conclusion}
We propose using generative AI to create a richly annotated medical-imaging dataset, and show that adding these synthetic images improves the performance of explainable classification models.
Our motivation is twofold: (1) incorporating domain knowledge, such as radiological diagnostic criteria, to enable human-like reasoning and foster trust in the model, and (2) overcoming the scarcity of annotated datasets that include such rich diagnostic metadata.

To this end, we propose an attribute-conditional diffusion model that employs a cross-attention mechanism to guide image generation based on non-binary radiological attributes. Applied to the LIDC-IDRI dataset, our model produces images with high subjective realism, as supported by a user study. While quantitative differences remain compared to real data, even a limited number of labeled real samples (for example, 20 real samples) was sufficient for the synthetic data to improve the performance of a downstream explainable classification model.

In particular, incorporating 2000 synthetic images improved attribute prediction accuracy by 13.4\% and target prediction accuracy by 1.8\% for the HierViT model, demonstrating the practical value of our method. Although a performance gap persists when compared to a real, fully annotated large-scale dataset (\#19713), such attribute-annotated datasets are difficult to obtain due to the substantial annotation effort required.

Our findings highlight the potential of domain-knowledge-driven generative AI to support the development of more accurate and interpretable classification systems in data-scarce medical contexts. Future work could explore extending this approach to other explainability modalities, such as generating paired textual explanations, and integrating them into multi-modal AI pipelines.

\begin{credits}
\subsubsection{\ackname} 
We thank all participants of the University Hospital Ulm for their contribution to the user study. 
This research was supported by the German Federal Ministry of Research, Technology and Space (BMFTR) within RACOON COMBINE "NUM 2.0" (FKZ: 01KX2121).
% A bold run-in heading in small font size at the end of the paper is
% used for general acknowledgments, for example: This study was funded

\subsubsection{\discintname}
The authors have no competing interests to declare that are relevant to the content of this article.
% It is now necessary to declare any competing interests or to specifically
% state that the authors have no competing interests. Please place the
% statement with a bold run-in heading in small font size beneath the
% (optional) acknowledgments\footnote{If EquinOCS, our proceedings submission
% system, is used, then the disclaimer can be provided directly in the system.},
% for example: The authors have no competing interests to declare that are
% relevant to the content of this article. Or: Author A has received research
% grants from Company W. Author B has received a speaker honorarium from
% Company X and owns stock in Company Y. Author C is a member of committee Z.
\end{credits}
\bibliographystyle{splncs04}

\end{document}